\documentclass[10pt,twocolumn,letterpaper]{article}

\usepackage{booktabs}
\usepackage{iccv}
\usepackage{times}
\usepackage{epsfig}
\usepackage{graphicx}
\usepackage{caption}
\usepackage{amsmath}
\usepackage{amssymb}
\usepackage{wrapfig}
\usepackage{arydshln}
\usepackage{subcaption}
\usepackage{xcolor}
\DeclareMathOperator*{\argmax}{arg\,max}

\usepackage[pagebackref=true,breaklinks=true,letterpaper=true,colorlinks,bookmarks=false]{hyperref}

\iccvfinalcopy 

\def\iccvPaperID{2289} 

\ificcvfinal\fi
\begin{document}

\title{Rescan: Inductive Instance Segmentation for Indoor RGBD Scans}

\author{
    \hfill
	Maciej Halber\qquad
	Yifei Shi\qquad
	Kai Xu\qquad
	Thomas Funkhouser
	\hfill
	\vspace{0.1cm}
	\\
	\hfill
	Princeton University
    \hfill
	\vspace{-1cm}
}

\twocolumn[{%
\renewcommand\twocolumn[1][]{#1}%
\maketitle
\begin{center}
    \centering
    \includegraphics[width=\textwidth]{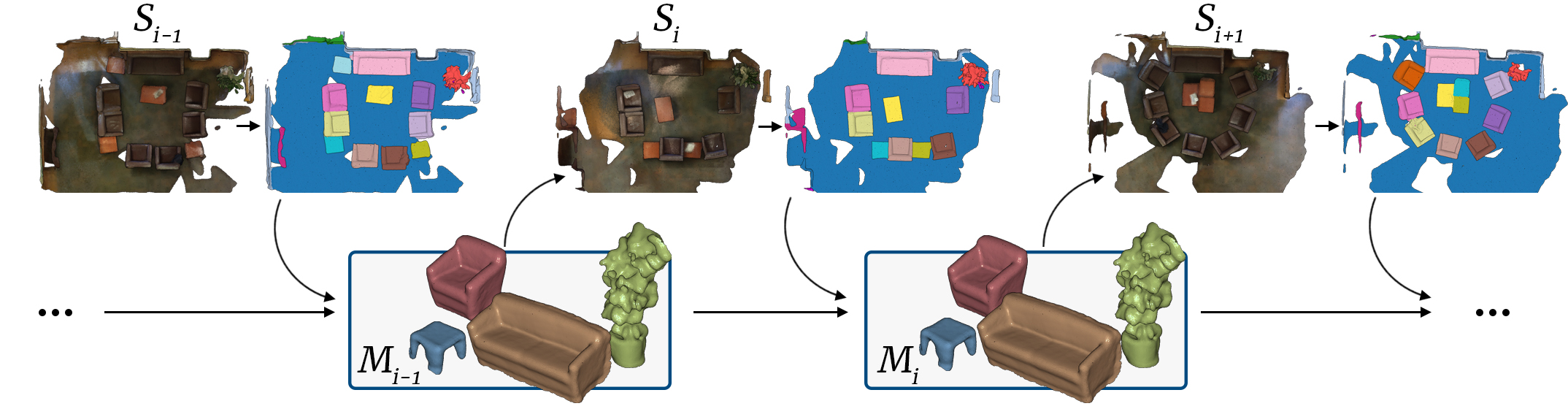}
    \captionof{figure}{ The proposed method estimates a persistent, temporally-aware scene model $M_i$ from a series of scene observations $S_i$, captured at sparse time intervals. $M_{i-1}$ is used to estimate an arrangement of objects in each novel observation $S_i$. The estimated arrangement is used to estimate the instance segmentation of $S_i$, which is then used to update the model $M_i$. }
    \label{fig:teaser}
\end{center}%
}]
\thispagestyle{empty}

\newcommand{\ignorethis } [1] {}
\newcommand{\redund}[1]{#1}

\newcommand{\chapnum    } [1] {\ref{#1}}
\newcommand{\appnum     } [1] {\ref{#1}}
\newcommand{\sectnum    } [1] {\ref{#1}}
\newcommand{\tblnum     } [1] {\ref{#1}}
\newcommand{\fignum     } [1] {\ref{#1}}
\newcommand{\eqnnum     } [1] {\mbox{(\ref{#1})}}
\newcommand{\chap       } [1] {Chapter~\chapnum{#1}}
\newcommand{\chaps      } [1] {Chapters~\chapnum{#1}}
\newcommand{\app        } [1] {Appendix~\appnum{#1}}
\newcommand{\apps       } [1] {Appendices~\appnum{#1}}
\newcommand{\sect       } [1] {Section~\sectnum{#1}}
\newcommand{\sects      } [1] {Sections~\sectnum{#1}}
\newcommand{\tbl        } [1] {Table~\tblnum{#1}}
\newcommand{\tbls       } [1] {Tables~\tblnum{#1}}
\newcommand{\fig        } [1] {Figure~\fignum{#1}}
\newcommand{\figs       } [1] {Figures~\fignum{#1}}
\newcommand{\eqn        } [1] {Equation~\eqnnum{#1}}
\newcommand{\eqns       } [1] {Equations~\eqnnum{#1}}

\newcommand{\apriori    }     {\textit{a~priori}}
\newcommand{\aposteriori}     {\textit{a~posteriori}}
\newcommand{\perse      }     {\textit{per~se}}
\newcommand{\naive      }     {{na\"{\i}ve}}

\newcommand{\Identity   }     {\mat{I}}
\newcommand{\Zero       }     {\mathbf{0}}
\newcommand{\Reals      }     {{\textrm{I\kern-0.18em R}}}
\newcommand{\isdefined  }     {\mbox{\hspace{0.5ex}:=\hspace{0.5ex}}}
\newcommand{\texthalf   }     {\ensuremath{\textstyle\frac{1}{2}}}
\newcommand{\half       }     {\ensuremath{\frac{1}{2}}}
\newcommand{\third      }     {\ensuremath{\frac{1}{3}}}
\newcommand{\fourth      }    {\ensuremath{\frac{1}{4}}}

\newcommand{\degree} {\ensuremath{^{\circ}}}

\renewcommand{\vec      } [1] {{\text{$\mathbf{#1}$}}}
\newcommand{\mat        } [1] {{\text{$\mathbf{#1}$}}}
\newcommand{\Approx     } [1] {\widetilde{#1}}
\newcommand{\change     } [1] {\mbox{{\footnotesize $\Delta$} \kern-3pt}#1}

\newcommand{\Order      } [1] {O(#1)}
\newcommand{\set        } [1] {{\lbrace #1 \rbrace}}
\newcommand{\floor      } [1] {{\lfloor #1 \rfloor}}
\newcommand{\ceil       } [1] {{\lceil  #1 \rceil }}
\newcommand{\inverse    } [1] {{#1}^{-1}}
\newcommand{\transpose  } [1] {{#1}^\mathrm{T}}
\newcommand{\invtransp  } [1] {{#1}^{-\mathrm{T}}}

\newcommand{\abs        } [1] {{| #1 |}}
\newcommand{\Abs        } [1] {{\left| #1 \right|}}
\newcommand{\norm       } [1] {{\| #1 \|}}
\newcommand{\Norm       } [1] {{\left\| #1 \right\|}}
\newcommand{\pnorm      } [2] {\norm{#1}_{#2}}
\newcommand{\Pnorm      } [2] {\Norm{#1}_{#2}}
\newcommand{\inner      } [2] {{\langle {#1} \, | \, {#2} \rangle}}
\newcommand{\Inner      } [2] {{\left\langle \begin{array}{@{}c|c@{}}
                               \displaystyle {#1} & \displaystyle {#2}
                               \end{array} \right\rangle}}


\pretolerance 800

\newlength{\w}
\newlength{\h}
\newlength{\x}

\definecolor{darkred}{rgb}{0.7,0.1,0.1}
\definecolor{darkgreen}{rgb}{0.1,0.7,0.1}
\definecolor{cyan}{rgb}{0.7,0.0,0.7}
\definecolor{dblue}{rgb}{0.2,0.2,0.8}
\definecolor{maroon}{rgb}{0.76,.13,.28}
\definecolor{burntorange}{rgb}{0.81,.33,0}

\ifdefined\ShowNotes
  \newcommand{\colornote}[3]{{\color{#1}\bf{#2: #3}\normalfont}}
\else
  \newcommand{\colornote}[3]{}
\fi
\newcommand{\ra}[1]{\renewcommand{\arraystretch}{#1}}

\newcommand {\todo}[1]{\colornote{cyan}{Note}{#1}}
\newcommand {\maciej}[1]{\colornote{burntorange}{MH}{#1}}
\newcommand {\tom}[1]{\colornote{blue}{TF}{#1}}
\newcommand {\kevin}[1]{\colornote{red}{KX}{#1}}
\newcommand {\yifei}[1]{\colornote{green}{YS}{#1}}
\newcommand{\mypara}{\par\vspace*{1mm}\noindent\textbf}

\definecolor{orange}{rgb}{1,0.5,0}
\newcommand {\revised}[1]{{#1}}
\newcommand {\revisedtable}{}

\newcommand\todosilent[1]{}

\ifdefined\SmallImages
  \newcommand{\images}{{images-small}}
\else
  \newcommand{\images}{{images}}
\fi

\newcommand{\iSet}{\ensuremath{\mathcal{I}}}
\newcommand{\wSet}{\ensuremath{\Omega}}
\newcommand{\grid}{\ensuremath{\mathcal{L}}}
\newcommand{\dMatrix}{\ensuremath{{D}}}
\newcommand{\scale}{\ensuremath{c}}
\newcommand{\ass}{\ensuremath{a}}
\newcommand{\perm}{\ensuremath{\pi}}
\newcommand{\iDist}{\ensuremath{d}}
\newcommand{\gDist}{\ensuremath{{d\,}'}}
\newcommand{\Ep}{\ensuremath{\mathcal{E}_p}}
\newcommand{\EpNorm}{\ensuremath{E_p}}
\newcommand{\Etwo}{\ensuremath{E_2}}
\newcommand{\Eone}{\ensuremath{E_1}}

\begin{abstract}
 In depth-sensing applications ranging from home robotics to AR/VR, it will be common to acquire 3D scans of interior spaces repeatedly at sparse time intervals (e.g., as part of regular daily use).  We propose an algorithm that analyzes these ``rescans'' to infer a temporal model of a scene with semantic instance information.   Our algorithm operates inductively by using the temporal model resulting from past observations to infer an instance segmentation of a new scan, which is then used to update the temporal model. The model contains object instance associations across time and thus can be used to track individual objects, even though there are only sparse observations.  During experiments with a new benchmark for the new task, our algorithm outperforms alternate approaches based on state-of-the-art networks for semantic instance segmentation.

\end{abstract}


\begin{figure*}
    \centering
    \includegraphics[width=\textwidth]{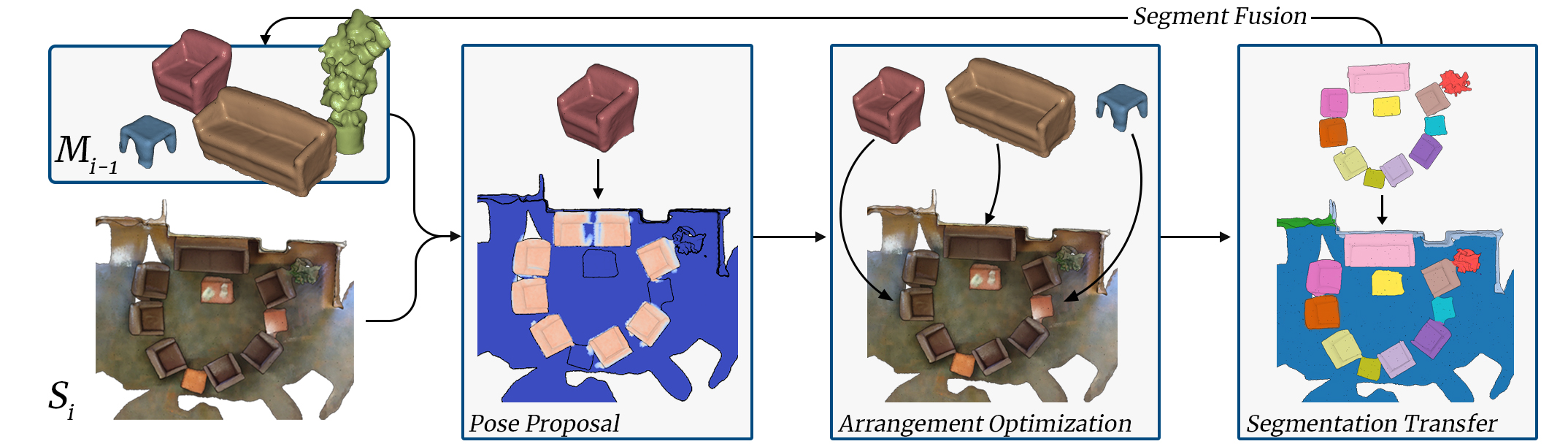}
    \captionof{figure}{A single inductive step of the proposed method. Given a novel scene observation $S_i$ and a model from the past $M_{i-1}$, our goal is to create an updated model $M_{i}$.   We first perform \textit{Pose Proposal}, where we search for a set of potential locations for each object in $M_{i-1}$ .   Then, we perform \textit{Arrangement Optimization}, where we search for the selection and arrangement of objects to minimize an objective function.   Then, we perform \textit{Segmentation Transfer}, in which $S_i$ is annotated with semantic instance labels from $M_{i-1}$. Finally, geometry from segments in $S_i$ is fused with $M_{i-1}$ to create an updated model ${M_i}$. }
    \label{fig:overview}
    \vspace*{-3mm}
\end{figure*}

\section{Introduction}
\label{sec:intro}

With the proliferation of RGBD cameras, 3D data is now more widely available than ever before \cite{dai2017scannet, scenenn-3dv16, Matterport3D}.   As depth capturing devices become smaller and more affordable, and as they operate in everyday applications (AR/VR, home robotics, autonomous navigation, etc.), it is plausible to expect that 3D scans of most environments will be acquired on a daily basis.   We can expect that 3D reconstructions of many spaces, visited at different times and captured from different viewpoints, will be available in the future, just like photographs are today.

In this paper, we investigate how repeated, infrequent scans captured with handheld RGBD cameras can be used to build a spatio-temporal model of an interior environment, complete with object instance semantics and associations across time.   The challenges are that: 1) each RGBD scan captures the environment from different viewpoints, possibly with noisy data; and 2) scans separated by long time intervals (once per day, every Tuesday, etc.) can have large differences due to object motion, entry, or removal.  Thus simple algorithms that perform object detection individually for each scan and/or simply cluster object detections and poses in space-time will not solve the problem.   Moreover, since large training sets are not available for this task, it is not practical to train a neural network to solve it.

We propose an inductive algorithm that infers information about new RGBD capture of a scene $S_i$ from a temporal model $M_{i-1}$ obtained from previous observations of $S$ (fig. \ref{fig:teaser}).   The input to the algorithm is the model $M_{i-1}$, representing all previous scans and a novel scene scan $S_i$. The output is an updated model $M_i$ that describes the set of objects $\mathcal{O}$ appearing in the scene and an arrangement $\mathcal{A}$ of those objects at each time step, including the most recent. 
At every iteration, our algorithm optimizes for the arrangement $A_i$ of objects in $S_i$, and then uses $A_i$ to infer the semantic instance segmentation of $S_i$. Segmentation of $S_i$ is then used to update object set $\mathcal{O}$ (see fig. \ref{fig:overview}).

To evaluate our algorithm we present a novel benchmark dataset that contains temporally consistent ground-truth semantic instance labels, describing object associations across time within each scene. Experiments with this benchmark suggest that our proposed optimization strategy is superior to alternative approaches based on deep learning for semantic and instance segmentation tasks.

Overall, the contributions of the paper are three-fold:
\begin{itemize}
    \setlength\itemsep{-0.5em}
    \item A system for building a spatio-temporal model for an indoor environment from infrequent scans acquired with hand-held RGBD cameras,
    \item An inductive algorithm that jointly infers the shapes, placements, and associations of objects from infrequent RGBD scans by utilizing data from past scans, 
    \item A benchmark dataset with rescans of 13 scenes acquired at 45 time-steps in total, along with ground-truth annotations for object instances and associations across time.
\end{itemize}

\section{Related Work}
\label{sec:related}


Most work in computer vision on RGBD scanning of dynamic scenes has focused on tracking \cite{song2013tracking} and reconstruction \cite{Newcombe15DynFusion}.  For example, Newcombe et al. \cite{Newcombe15DynFusion} showcases a system where multiple observations of a deforming object are fused into a single consistent reconstruction.  Yan et al. \cite{Yan14ProactiveScanning} scan moving articulated shapes by tracking parts as they are deformed over time.  These methods differ from ours as they require observation of motions as they occur.

For sparse temporal observations, early work in robotics focuses on the analysis of 2D maps created from 1D laser range sensors \cite{anguelov2002learning,biswas2002towards,gallagher2009gatmo}.  For example, Biswas \cite{biswas2002towards} used 1D laser data to detect objects within a scene and associate them across time. However, their method relies upon 2D algorithms and assumes that object instances cannot overlap across time, which makes it inapplicable in our setting. More recently, image based techniques for sparse observations were proposed --- Shin \cite{Shin13} extends SfM to also predict poses of moving objects. 

Other work has aimed at life-long scene understanding using data captured with actively controlled sensors \cite{faulhammer2017autonomous,krajnik2017fremen,santos2017spatio,young2017making}.   For example, several algorithms proposed in the STRANDS project \cite{hawes2017strands} process the scenes observed from a repeated set of views \cite{Ambrus14Metarooms,bore2017detection,schulz2001probabilistic}.  Others focus on controlling camera trajectories to acquire the best views for object modeling \cite{ekekrantztowards,faulhammer2017autonomous} and/or change detection \cite{ambrus2015unsupervised}.   These problems are different than ours, as we focus on analyzing previously acquired RGBD data captured without a specifically tailored robotic platform and active control. 

Some work in computer vision has focused on change detection and segmentation of dynamic objects in RGBD scans \cite{fehr2017tsdf,Lee17STSLAM,wang2003online}.  For example, Fehr et al. \cite{fehr2017tsdf} showcases a system for using multiple scene observations to classify surface elements as dynamic or static.  Wang et al.\cite{wang2002simultaneous} detect moving objects so that they can be removed from a SLAM optimization.  Lee et al. \cite{Lee17STSLAM} propose a probabilistic model to isolate temporally varying surface patches to improve camera localization.   While operating on RGBD captures from handheld devices, these methods do not produce instance-level semantic segmentations, nor do they generate associations between objects across time.
 
More recent work has focused on automatic clustering of 3D points into clusters across space and time \cite{finman2014physical,herbst2014toward}.  For example, Herbst et al. \cite{herbst2014toward} jointly segments multiple RGBD scans with a joint MRF formulation.   Finman et al. \cite{finman2014physical} detects clusters of points from pairwise scene differencing and associates new detections with previous observations.  Although similar in spirit to our formulation, these methods operate only on clusters of points, without semantics, and thus are not suited for applications that require semantic understanding of how objects move across space-time. 

Finally, many projects have considered temporal modeling of environments in specific application domains.   For example, several systems in civil engineering track changes to a Building Information Model (BIM) by alignment to 3D scans acquired at sparse temporal intervals \cite{golparvar2012automated,karsch2014constructaide,rebolj2017point,tuttas2017acquisition}. They generally start with a specific building design model \cite{han2015bim}, construction schedule \cite{turkan2012automated}, and/or object-level CAD models \cite{bosche2009automated}, and thus are not as general as our approach.   The Scene Chronology project \cite{Matzen14SceneChronology} and others \cite{martin2015time,Schindler07Cities} build temporal models of cities from image collections -- however, they do not recover a full 3D model with temporal associations of object instances as we do. 

\section{Algorithm}
\label{sec:algorithm}

\subsection{Scene Representation}


Our system represents a scene at time $t_i$ with a temporal model $M_i$ comprising a tuple $\{\mathcal{O}, \mathcal{A}\}$, where $\mathcal{O} = \{o_0, \ldots, o_n\}$ is a list of $n$ object instances that have appeared within this or any prior observation $S_j$ for $j\in[0,i]$, and $\mathcal{A} = \{A_0, \ldots, A_i\}$ is a list of object arrangements estimated for each observation $S_j$.   Each object instance $o_k$ is represented by $\{u_k, G_k, c_k\}$, where $u_k$ is unique instance id, $G_k$ is the object's geometry, and $c_k$ is the semantic class. Each arrangement $A_i$ is a list of poses $\{a_i^0, \ldots, a_i^m\}$, where $a_i^j = \{u_j, \mathbf{T}_j, s_j\}$. $u_j$ is the unique id of $j$-th object and function $\Omega(u_j)$ returns index $k$ to $\mathcal{O}$. $\mathbf{T}_j$ is a transformation that moves geometry $G_k$ into correct location within the scene $S_i$. Lastly $s_j$ is a matching score quantifying how well $\mathbf{T}_jG_k$ matches the geometry of $S_i$.

\subsection{Overview}

Our algorithm updates the temporal model in an inductive fashion.  Given the previous model $M_{i-1}$ and a new scan $S_i$, we predict a new model $M_i$ (see fig. \ref{fig:overview}) by executing four consecutive steps.   The first proposes potential poses for objects in $\mathcal{O}$ (sec. \ref{sec:poseproposal}).   The second performs a combinatorial optimization to find the arrangement $A_i$ that maximizes a new objective function jointly accounting for geometric fit and temporal coherence (sec. \ref{sec:arrangementoptimization}). The third step uses $\mathcal{O}$ and $A_i$ to infer an instance-level semantic segmentation of $S_i$.   The fourth step updates the geometry $G_k$ of each object $\in A_i$ by aggregating its respective segment from $S_i$.   The following four subsections offer the details on how each of these steps is implemented.

\subsection{Object Pose Proposal}
\label{sec:poseproposal}

The first step of our pipeline is to find a set of potential placements for each object $o_k \in \mathcal{O}$, creating a search space for the Arrangement Optimization stage (sec. \ref{sec:arrangementoptimization}).  Formally, the input to this stage is a set of objects $\mathcal{O}$ and a scan $S_i$. The output is a set $\mathcal{P}$ of scored pose lists $P_k = \{p_k^0, \ldots, p_k^x\}$ for each object $o_k$.   A scored pose $p_k^l$ is a tuple $\{\mathbf{T}_k^l, s_k^l\}$, where $\mathbf{T}_k^l$ is the proposed rigid-body transformation and $s_k^l$ is a geometric matching score describing how well pose $\mathbf{T}_k^l$ aligns $G_k$ to the geometry of $S_i$. 


Finding transformations that align surfaces $A$ and $B$ is a longstanding problem in computer graphics and vision \cite{Rusinkiewicz01ICP}.   In our setting, we wish to find a set of poses for the surface $A$ with good alignment with surface $B$, where $A=o_k$ and $B=S_i$.   Prior work usually attempts to solve similar problems by employing feature-based methods.   Such methods sub-sample the two surfaces to obtain a set of meaningful keypoints and then match them to produce a plausible pose (e.g., using Point-Pair Feature matching\cite{Drost10PPF}). However, as it has been noted in other domains, keypoints may limit the amount of information a method considers, with dense matching methods leading to less failures \cite{engel14eccv}. 

Following this intuition, we propose a dense matching procedure, where we slide each of the objects $o_k$ across the scene, perform an ICP optimization at each of the discrete locations and compute a matching score based on the traditional point-to-plane distance metric \cite{low2004linear}. 

This approach might seem counter-intuitive, as a naive implementation of such grid-search would lead to a prohibitive run-time performance.   We find however that such an approach can be made acceptably fast while leading to much better recovery of correct poses.   To speed-up the run-time performance of our method we make use of the multi-resolution approach. We compute a four-level hierarchy for the input point cloud (the geometries $G_k$), with minimum distance between any two points at a level equal to $\{0.01m, 0.02m, 0.04m, 0.08m\}$ respectively.   To compute this representation we follow an algorithm described in \cite{Corsini11BlueNoiseSurface}. Multi-resolution representation allows us to perform the dense search only on the coarsest level of the hierarchy, and return a subset of poses with sufficiently high scores to be verified at higher levels, leading to significant performance gains.   Additionally, we make a simplifying, but reasonable assumption that objects in our scenes move on the ground plane and rotate around the gravity direction. 

\begin{figure}[tb]
    \centering
    \includegraphics[width=0.5\textwidth]{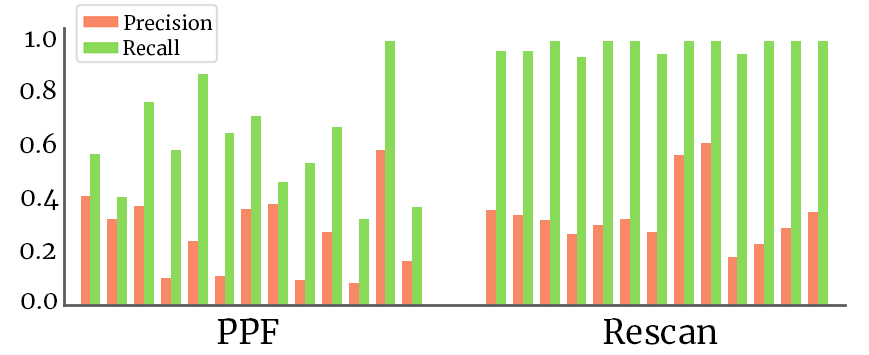}
    \caption{Comparison of the precision/recall scores obtained for all scenes in our database, comparing PPF matching \cite{Birdal15PPF} to our method. In our experiments a pose of an object $o_k$ is considered a true positive if the distance between object centers is less than $0.2m$ and object's classes agree.}
    \label{fig:ppfdensecomp}
\end{figure}

With this approach we are able to produce a set $\mathcal{P}$ of pose lists $P_k$ for each object $o_k$ in $\mathcal{O}$. The advantage of this dense grid-search method is that it produces sets of poses that contain most of the true candidate locations, even if the local geometry of $S_i$ might be different from $G_k$ due to reconstruction errors. We showcase the comparison to keypoint based methods \cite{Drost10PPF, Birdal15PPF} in figure \ref{fig:ppfdensecomp}.

\subsection{Arrangement Optimization}
\label{sec:arrangementoptimization}

In the second step our algorithm selects a subset of poses from the previous step to form an object arrangement.  The input is a set of objects $\mathcal{O}$, a set of pose lists $\mathcal{P} = \{P_0,\ldots,P_k\}$ for each object $o_k$, and the scan $S_i$. The output is an arrangement $A_i$ that describes a global configuration of objects which maximizes the objective. 

This problem statement leads to a discrete, combinatorial optimization.  First reason for choosing this approach is that the number of objects within the scene $S_i$ is not known a priori. A combinatorial approach allows us to propose arrangements $A_i$ of variable lengths, that will adapt to the contents of $S_i$. A second reason is that finding the optimum requires global optimization -- the placement of one object can greatly affect the placement of another.  Additionally, deep learning is hard to apply in this instance due to the lack of the training data, as well as the non-linearity of the proposed objective function.

\subsubsection{Objective Function}
\label{sec:objectivefunction}
To quantify the quality of the candidate arrangement $A'_i$ we use the objective function that is a linear combination of the following four terms:
\begin{align*}
    O(S_i, A'_i, \mathcal{A}) & = w_cO_c(S_i, A'_i) && \text{Coverage Term} \\
            & + w_gO_g(S_i, A'_i) && \text{Geometry Term}\\
            & + w_iO_r(A'_i) && \text{Intersection Term}\\ 
            & + w_hO_h(A'_i,\mathcal{A}) && \text{Hysteresis Term}\\ 
\end{align*}
Each term $O_x$ produces a scalar value $\in[0, 1]$ that describes the quality of $A'_i$ w.r.t. that specific term. We use grid search to find good values for the weights $\vec{w}=\{2.0, 0.3, 1.0, 1.8\}$, which express the relative importance of each term. 

\setlength{\columnsep}{5pt}%
\setlength{\intextsep}{0pt}%

\begin{wrapfigure}{r}{0.19\textwidth}
    \centering
    \includegraphics[width=0.19\textwidth]{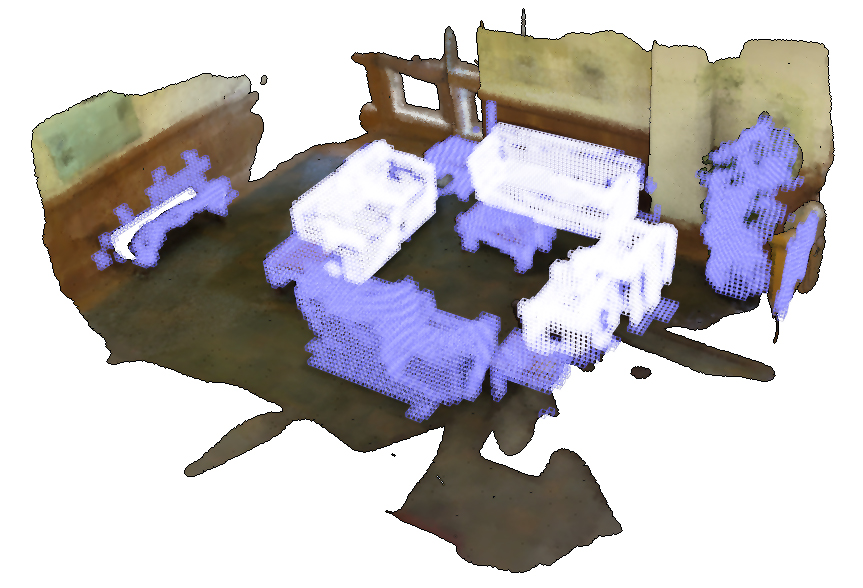}
\end{wrapfigure}

\noindent\textbf{The Coverage term} measures the percentage of the scene that is covered by objects in $A'_i$. The intuition behind this term is that every part of the scene should ideally be explained by some object in $A'_i$. $O_c(S_i, A'_i)$ takes as input a scene $S_i$ and the candidate arrangement $A'_i$. To compute $O_c(S_i, A'_i)$ we voxelize both the scene $S_i$ and the objects in $A'_i$, resulting in two 3D grids $V_S$ and $V_A$. The $O_c(S_i, A'_i)$ is calculated as the number of cells that are equal in both grids, over the number of cells in $V_S$ - $O_c(S_i, A'_i) = \frac{|V_s(j) \land V_A(j)|}{|V_s(j)|}$. For this formula to be accurate we need to ensure however that we only voxelize the dynamic parts of the scene $S_i$. As such we deactivate any cells in $V_S$ that belong to the static parts of the scene, like walls and floor, which can easily be detected with a method like RANSAC \cite{Fischler81RANSAC}. The inset figure above showcases a visualization of both grids $V_S$ (blue cells) and $V_A$ (white cells). As seen there, the $V_S$ covers the non-static parts of the scene only, leading to $O_c$ being a good estimate of the coverage.

\noindent\textbf{The Geometry term} is a measure of the geometrical agreement between the scene $S_i$ and objects in the candidate arrangement $A'_i$. We include this term to guide the objective function to select objects that best match the geometry of the scene at a specific location. This value is simply computed as an average of scores $s_k^l$ from the procedure described in section \ref{sec:poseproposal}. $O_g(S_i, A'_i) = \frac{\sum_k g(a_i^j)}{|A'_i|}$, where $g(a_i^j)$ returns the geometrical score fit for placement of object $o_j$.

\begin{wrapfigure}{r}{0.2\textwidth}
    \centering
    \includegraphics[width=0.2\textwidth]{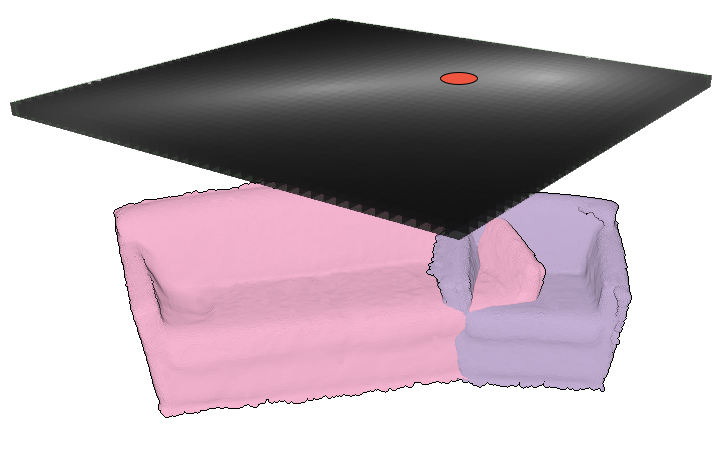}
\end{wrapfigure}

\noindent\textbf{The Intersection term} aims to estimate how much a pair of objects in the arrangement $A'_i$ interpenetrate. Intuitively, such interpenetration would mean that two objects occupy the same physical location, which implies an impossible configuration. 
In our approach, we compute a coarse approximation of this term. First, we compute a covariance matrix $\mathbf{\Sigma}_k$ of each $G_k$. Covariances for each object allow us to compute a symmetric Mahalanobis distance $SD_M$ between objects to approximately quantify how close they are to each other. $SD_M(O_r, o_j) = 0.5(D_M(m_{ij}, \mathbf{T}_ic_i, \Sigma_i) + D_M( m_{ij}, \mathbf{T}_jc_j, \Sigma_j))$, where $\mathbf{T}_ic_i$,$\mathbf{T}_jc_j$ are transformed centroids of $G_i$, $G_k$, the midpoint between them is $m_{ij}$, and function $D_M$ is the Mahalanobis distance. With $SD_M$ computed for all pairs of objects $o_k$, the value $O_r(A'_i)$ is $1-||\{\exp(\frac{-SD^2_M(o_0, o_1)}{2\sigma^2}), \ldots, \exp(\frac{-SD^2_M(o_{n-1}, o_n)}{2\sigma^2})\}||_\infty$. The rationale behind the use of the infinity norm is to generate a high penalty if just a single pair of objects exhibits a low score interpenetration. The inset figure above showcases a visualization of $SD_M$ for two intersecting objects. The point at which we evaluate the $SD_M$ is marked with red, showcasing high values in regions where either or both objects are present, and low values in the free space. It is also clear that the value of $SD_M$ would be higher if the objects interpenetrated even more.

\begin{wrapfigure}{r}{0.15\textwidth}
    \centering
    \includegraphics[width=0.15\textwidth]{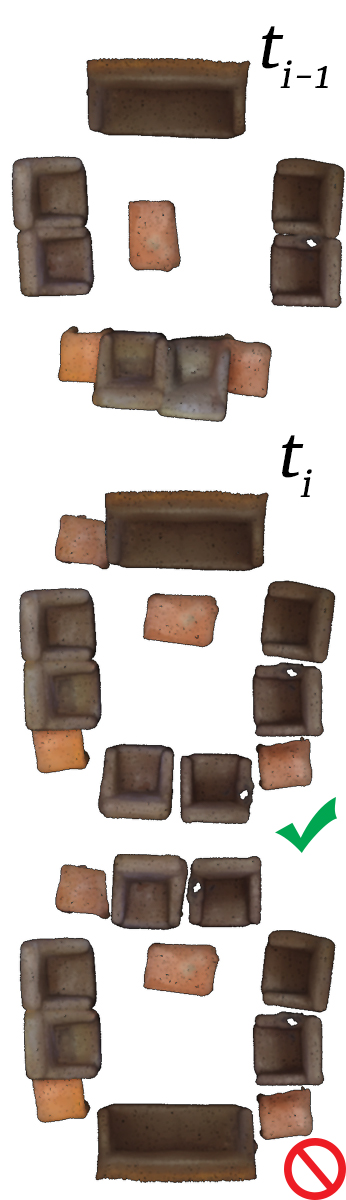}
\end{wrapfigure}

\noindent\textbf{The Hysteresis term} informs how well the current arrangement estimate $A'_i$ resembles a previously observed arrangements from the set $\mathcal{A}$. In addition it expresses our preference for a minimal relative motion. Each object in $A'_i$ is assigned a score, with the value based on whether $u_k$ is a novel instance, or has been observed in the past. In the former case, we assign a novel object constant score $h=0.4$ (found manually). In the latter, the score is $h + (1-h)exp(\frac{-||T(c_k, i) - T(c_k, j)||_2}{2\sigma^2})$. $T(c_l,j)$ is a function that applies the appropriate transformation to centroid $c_l$ at time $t_j$. As a result, novel objects will be always preferred, unless they have undergone a significant transformation.   In such a case, we would like $O_h$ to express that novel object appearances have similar probability. The value of $O_h(A_i,\mathcal{A}) $ is computed as an average of the above scores. The inset figure above illustrates an arrangement at $t_{i-1}$ and two possible arrangement estimates at $t_i$. The form of $O_h(A'_i,\mathcal{A})$ encourages the selection of middle arrangement as it does not contain significant motion the sofa and chairs.

\subsubsection{Optimization}

To find arrangement $A_i = \argmax_{A'_i}O(S_i, A'_i, \mathcal{A})$, we employ a combination of greedy initialization and simulated annealing. We begin by greedily selecting an object $o_k$ at a pose $p_k^l$ which improves objective the most. This process of addition is continued until the objective function starts decreasing. After this stage, we perform simulated annealing optimization. We run the simulated annealing for 25k iterations, using a linear cooling schedule with a random restarts ($0.5\%$ probability to return to the best scoring state). To explore the search space we use the following actions with a randomly selected object $o_k$:
\vspace*{-1mm}
\begin{itemize}
    \setlength\itemsep{-0.25em}
    \item \textbf{Add Object} - We add $o_k$ at a random pose $p_k^l$ to ${A'_i}$.
    \item \textbf{Remove Object} - We remove $o_k$ from ${A'_i}$.
    \item \textbf{Move Object} - We select $o_k$ from ${A'_i}$ and assign it new pose $p_k^m$.
    \item \textbf{Swap Objects} - We swap the location of $o_k$ and $o_l$, another randomly selected object of the same semantic class.
\end{itemize}

\subsection{Segmentation Transfer}
\label{sec:segmenttransfer}

The third step of the algorithm transfers the semantic and instance labels from $A_i$ to scan $S_i$.  The estimated arrangement from the previous step can be used to perform segmentation transfer, as we have semantic class $c_k$ and instance id $u_k$ associated with each object in $\mathcal{O}$.  Using the estimated pose $p_k^l$ for each of the objects $o_k$ in $A_i$, we transform its geometry $G_k$ to align with $S_i$.  We then perform a nearest neighbor lookup (with a maximum threshold $d = 5cm$ to account for outliers) and use the associations to copy both the instance and semantic labels from objects in $A_i$ to $S_i$. Since there is no guarantee that all points in $S_i$ will have a neighbor within the threshold $d$, we follow-up the lookup with label smoothing based on multi-label graph-cut \cite{Delong2012MultiLabel}.

\subsection{Geometry Fusion}
\label{sec:geometryfusion}

The final step of the algorithm is to update the object geometries $G_k$ for objects in $\mathcal{O}$.  To do so for each object $o_k \in A_i$, we extract the sub point clouds from $S_i$ that were assigned instance label $u_k$ in the previous step, and then we concatenate them with $G_k$ to generate new point cloud $G'_k$. In the idealized case, the two surfaces would be identical, as they represent the same object. However, due to partial observation, reconstruction, and alignment errors, we cannot expect that in practice. As such, we solve for a mean surface $\tilde{G}_k$ that minimizes the distance to all points in the $G'_k$, using Poisson Surface Reconstruction \cite{Kazhdan06PSR}. After this process, we uniformly sample points on the resulting surface $\tilde{G}_k$ to get a new estimate of  $G_k$ that will be used for matching when a new scene $S_{i+1}$ needs to be processed.

\section{Evaluation}
\label{sec:evaluation}

\begin{figure*}[!ht]
    \centering
    \includegraphics[width=\textwidth]{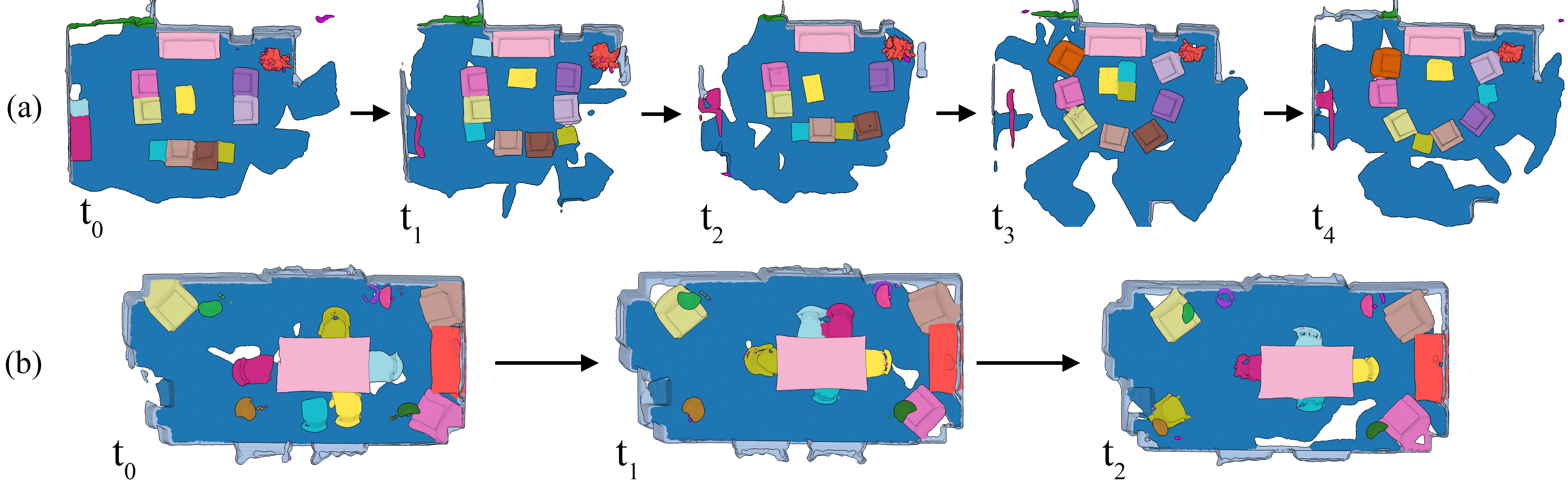}
    \captionof{figure}{Inductive instance segmentation results. Given a segmentation at time $t_0$, our method is able to iteratively transfer instance labels to future times, even when the number of the objects in the scene changes.}
    \label{fig:transferresult}
    \vspace*{-3mm}
\end{figure*}

Evaluation of the proposed algorithm is not straightforward, as there is little to no prior work directly addressing instance segmentation transfer between 3D scans.  

\vspace*{2mm}\noindent{\bf Dataset:} To evaluate the proposed approach, we have created a dataset of temporally varying scenes. Our dataset contains 13 distinct scenes, with total of 45 separate reconstructions.  Each scene contains between 3 to 5 scans, where objects within each catpure were moved to simulate changes occuring across long time periods.   Along with the captured data, we also provide manually-curated semantic category and instance labels for every object in every scene.  The instance labels are stable across time, providing associations between object instances in different scans, which we can use to evaluate our algorithms.  Additionally, we provide permutations of instance assignments for each scene to account for cases where objects' motion is ambiguous and multiple arrangements can be considered correct. More details about the dataset are included in the supplemental material. 

\vspace*{2mm}\noindent{\bf Metrics:} We evaluate our approach using three metrics.   The first is the \textit{Semantic Label} metric that measures the correctness of class labels -- it is implemented in the same way as the semantic segmentation task in the ScanNet Benchmark \cite{dai2017scannet} and is reported as mean class IoU.   The second is the \textit{Semantic Instance} metric that measures the correctness of the object instance separations -- it again comes from the ScanNet Benchmark \cite{dai2017scannet} and is reported as mean Average Precision (IoU=0.5). Third, we propose a novel \textit{Instance Transfer} metric, which specifically requires an agreement of instance indices across time. This metric is reported as mean IoU, where we count the number of points in both ground truth and prediction that share equivalent instance id. The \textit{Instance Transfer} metric is much more challenging, as it requires associating objects with specific instance ids in different scans.

\vspace*{2mm}\noindent{\bf Baseline:}  Given the success of the recent deep models for the scene understanding (as shown on the leaderboard of \cite{dai2017scannet}), it is interesting to compare the results of our algorithm to the best available method based on deep neural networks.   One of the best available methods for 3D instance segmentation is MASC \cite{Liu2019MASCMA}, which is based on semantic segmentation with SparseConvNet \cite{3DSemanticSegmentationWithSubmanifoldSparseConvNet}.   To test these methods on our tasks, we pre-trained the SparseConvNet and MASC models on ScanNet's training set. We performed fine-tuning of MASC with the ground-truth labels of first observation (time $t_0$) of each scene $S_0$ in our database. This fine-tuned model provides instance segmentation, which can be combined with the Hungarian method \cite{kuhn1955hungarian} to estimate instance associations across time.   This sequence of steps provides a very strong baseline combining state-of-the-art methods for instance segmentation with an established algorithm for assignment.

\subsection{Quantitative Results}
\begin{table}
\ra{1.1}
    \begin{tabular}{@{}lllll@{}} \toprule
         Method & \begin{tabular}[c]{@{}l@{}}Semantic \\ Label\end{tabular} & \begin{tabular}[c]{@{}l@{}}Semantic \\ Instance\end{tabular} & \begin{tabular}[c]{@{}l@{}}Instance\\ Transfer\end{tabular} \\ \midrule
        SparseConvNet      &   0.203   &    \quad-        &  \quad-  \\ 
        MASC               &   0.310   &    0.291    &  0.175  \\ 
        MASC (fine-tuned)\quad\quad    &   0.737   &    0.562    &  0.345  \\ \addlinespace[3pt]\hdashline[.4pt/1pt]\addlinespace[3pt]
        Rescan             &   0.859   &    0.837    &  0.650  \\ \bottomrule
    \end{tabular}
    \caption{Comparison of our method to SparseConvNet \cite{3DSemanticSegmentationWithSubmanifoldSparseConvNet} and MASC \cite{Liu2019MASCMA}. SparseConvNet does not produce instance labels, hence we omit reporting on the \textit{Semantic Instance} and \textit{Instance Transfer} task, and only fine-tune MASC. }
    \label{tbl:segtable}
    \vspace*{-3mm}
\end{table}

\begin{figure*}[!ht]
    \centering
    \includegraphics[width=\textwidth]{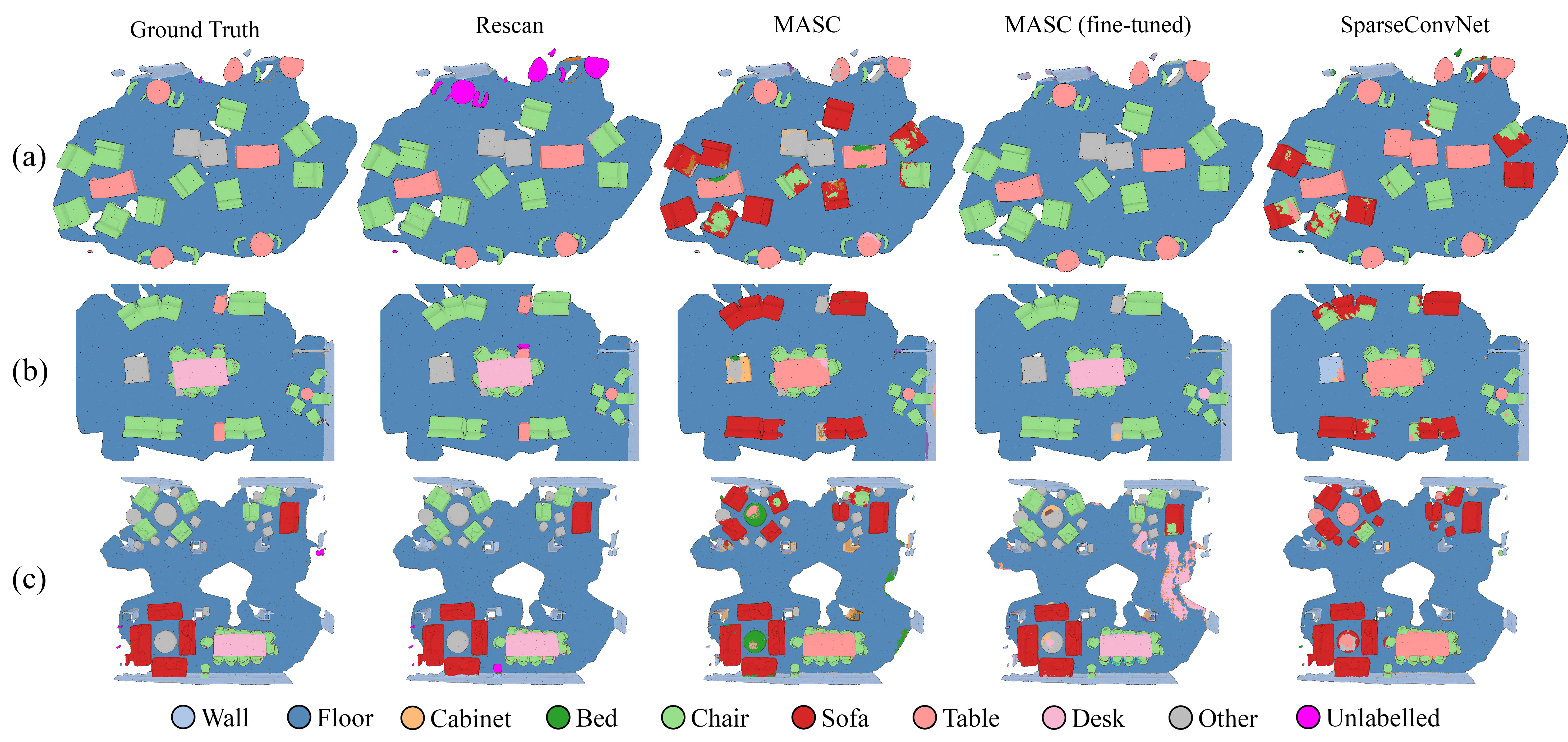}
    \captionof{figure}{ Qualitative comparison on the semantic segmentation task. Proposed method is able to provide high quality semantic labels as a result of instance segmentation transfer. Compared to competing methods, ours is able to produce better per object labels and does not confuse object classes. } 
    \label{fig:semanticresults}
    \vspace*{-3mm}
\end{figure*}

\vspace*{1mm}\noindent{\bf Evaluation and comparison:}  Since we solve an inductive task (predict the answer at $t_i$, given an answer at $t_{i-1}$), it is not obvious how to initialize the system for our experiments.  As our aim is to evaluate the inductive step alone, we chose to initialize time $t_0$ with a correct instance segmentation.  That choice avoids confounding problems with de novo instance segmentation at $t_0$ with the main objective of the experiment.  We have each algorithm in the experiment transfer the instance segmentation from $t_0$ to $t_1$, then transfer the result to $t_2$, and so on.

We ran this experiment for our method in direct comparison to the baseline.   Results for all three evaluation metrics are shown in Table \ref{tbl:segtable}.   They show that our algorithm significantly outperforms competing methods.   As expected, we see that the deep neural networks trained on the ScanNet training set \cite{dai2017scannet} do not perform very well on our data without fine-tuning.    After fine-tuning on the data in $S_0$, they do much better. Fine-tuning allows for a fair comparison, as both their and our methods have access to the same information from $S_0$ to predict labels for $S_i; i>0$.   Despite this, instance segmentation on later time steps still performs worse than our algorithm, and instance associations across time are poor.  We attribute the difference to the fact that our method is instance-centric, where the segmentation is inferred from the estimated objects' arrangement. This is in stark opposition to methods like MASC, where the instances are inferred from a semantic segmentation. 

\vspace*{1mm}\noindent{\bf Ablation studies:} Second, we present the results of ablation studies that showcase the influence of various terms in our objective function on the results in a specific task. As seen in table \ref{tab:ablationtable}, by far the most important term of our proposed objective is the \textit{Coverage Term}. Without it, the objective function is discouraged from adding more objects. The optimization simply finishes with a single object added to the scene - as adding any more would lead to a decrease in other terms. 

The second most important term, especially for the \textit{Instance Transfer} task, is the \textit{Hysteresis Term}. It is intuitive that lacking this term, the objective function is not encouraged to find an arrangement that will be consistent with previous object configurations. We note that when omitting this term, the semantic segmentation task achieves a slightly better result.   The reason is that to prevent addition of superfluous objects the novel objects are assigned relatively low score (sec. \ref{sec:objectivefunction}). Without the \textit{Hysteresis Term}, the proposed objective is free to insert additional objects - however their configuration is often not correct, leading to lower scores for other two tasks.   This result suggests that there exists a better formulation of the hysteresis function - an interesting direction for future research.   

The presence of the \textit{Intersection Term} is important for the \textit{Semantic Instance} and \textit{Instance Transfer} tasks. Intuitively, the semantic segmentation score is unaffected as it is often the case that intersecting objects share the semantic class. The \textit{Geometry Term} has the least influence on the results.   This is not surprising, as the poses that survived the pose proposal stage (see sec. \ref{sec:poseproposal}) were high scoring ones.

\subsection{Qualitative Results}
\begin{table}[t]
\ra{1.1}
    \centering
    \begin{tabular}{@{}lllll@{}} \toprule
     Method & \begin{tabular}[c]{@{}l@{}}Semantic \\ Label\end{tabular} & \begin{tabular}[c]{@{}l@{}}Semantic \\ Instance\end{tabular} & \begin{tabular}[c]{@{}l@{}}Instance \\ Transfer\end{tabular} \\ \midrule
    No Coverage Term       &   0.061   &    0.058    &  0.048  \\ 
    No Geometry Term       &   0.853   &    0.825    &  0.617   \\ 
    No Intersection Term   &   0.859   &    0.781    &  0.584  \\ 
    No Hysteresis Term     &   0.870   &    0.818    &  0.226  \\ \addlinespace[3pt]\hdashline[.4pt/1pt]\addlinespace[3pt]
    Full Method            &   0.859   &    0.837    &  0.650  \\ \bottomrule
    \end{tabular}
    \caption{Ablation study showcasing the influence of objective function terms on each of the proposed tasks.\vspace*{-2mm}}
    \label{tab:ablationtable}
\end{table}



\noindent{\bf Inductive segmentation transfer:} We showcase qualitative results for the \textit{Instance Transfer} task using our method in figure \ref{fig:transferresult}. Again, in this task we use the ground-truth segmentation provided by the user at $t_0$ and transfer it to all other observations sequentially. The results of such segmentation transfer offer stable and well-localized instances. Even over multiple time-steps, our method is able to keep track of objects identities, providing us with information on their location and motion. Additionally, thanks to the fact that the objective function prefers minimal change, we are able to deal with challenging configurations. For example in \ref{fig:transferresult}\textcolor{red}{a} our method is able to correctly recover three coffee tables at time $t_3$, despite their proximity and visual similarity.

\vspace*{1mm}\noindent{\bf Semantic segmentation:} Figure \ref{fig:semanticresults} showcases qualitative comparisons between our method and DNN-based methods \cite{Liu2019MASCMA, 3DSemanticSegmentationWithSubmanifoldSparseConvNet}. Without fine-tuning, the segmentation issues are obvious. Learned methods confuse labels like \textit{sofa} and \textit{chair}, which explains low scores in table \ref{tbl:segtable}. Fine-tuning helps reduce these effects - however we also see some overfitting errors. Our method is able to recover high quality semantic segmentation, where due to the fact that our approach is instance-centric, a single instance can not have more than a single semantic class. Our method's success is however dependent on the overlap between current and previous observations of $S$.   When lots of novel objects appear, the \textit{Hysteresis Term} might discouraging addition of all of them, as it aims to produce arrangement similar to previously observed ones (fig. \ref{fig:semanticresults}\textcolor{red}{a}).

\begin{figure}
    \centering
    \includegraphics[width=0.5\textwidth]{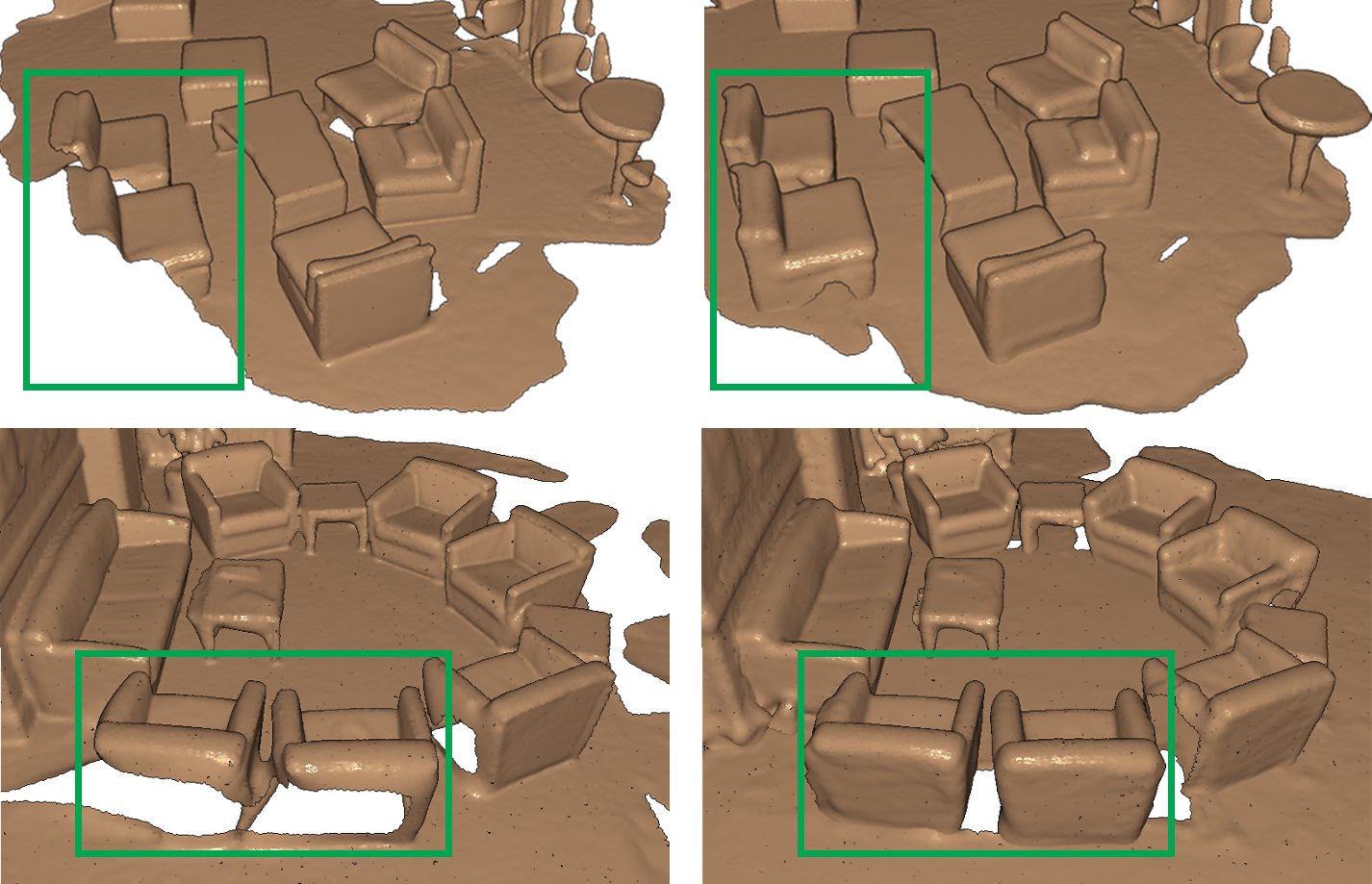}
    \captionof{figure}{Model completion results. The left column shows two scans of a scene with moving objects.  The right column shows our reconstruction of the scene using objects and locations from the temporal model $M$.}
    \label{fig:completion}
    \vspace*{-4mm}
\end{figure}

\vspace*{1mm}\noindent{\bf Model completion results:} Our method for aggregating the observations of moving objects from multiple time steps allows it to produce more complete surface reconstructions than would be possible otherwise.   Many other systems explicitly remove moving objects before creating a surface model (to avoid ghosting) \cite{Keller13PBF}.  Our approach uses the estimated object segmentations and transformations to aggregate points associated with each object $o_k$ to form a $G_k$ that is generally more complete than could be obtained from any one scan.  Composing the aggregated $G_k$ using transformations $T_k$ in each object arrangement $A_i$ provides a model completion result (fig. \ref{fig:completion}). 

\begin{figure}[t]
    \centering
    \includegraphics[width=0.5\textwidth]{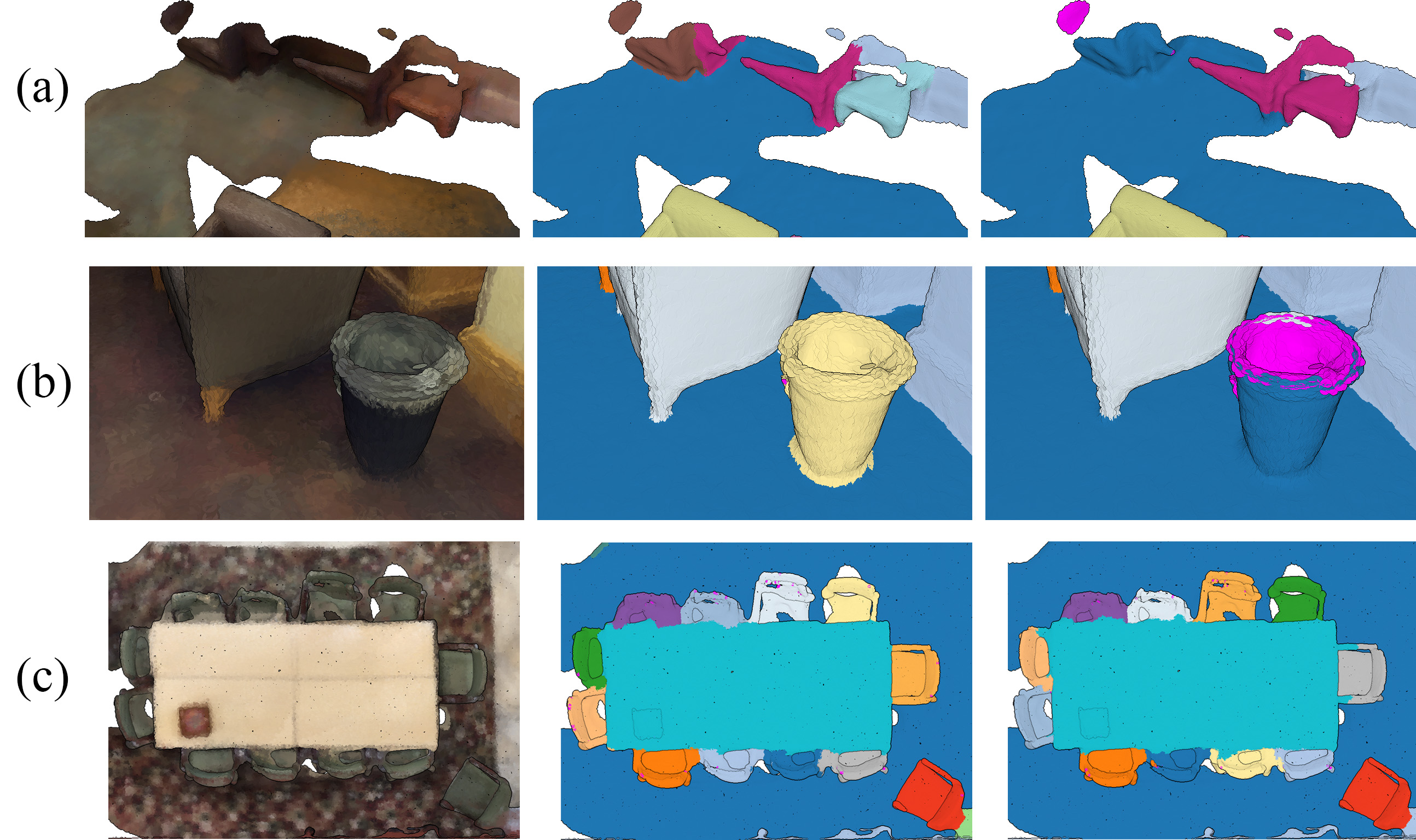}
    \captionof{figure}{Failure modes of the proposed method. (a) Partial scanning prevents the pose proposal stage from generating plausible poses. (b) Small objects contribute little to the coverage term. If such objects undergo significant motion our algorithm might miss them. (c) When similar, partially scanned objects are considered, our method might not produce the correct permutation. }
    \label{fig:failures}
    \vspace*{-4mm}
\end{figure}

\vspace*{1mm}\noindent{\bf Failures:} We identify three main failure modes of our approach (fig. \ref{fig:failures}).  The first issue arises due to the geometry focused nature of our approach. If the objects are only partially scanned, the pose proposal stage will not be able to recover highly scored poses. As such, these objects will simply not be added to the space of possible configurations that the optimization can choose from.  The second is caused by the limited contribution of small objects to the scene coverage score. Combined with a small \textit{Hysteresis Term} value under significant motion, the objective function might prefer not adding these objects.  Lastly, in cases like the one in figure \ref{fig:failures}\textcolor{red}{c}, an incorrect permutation of objects might have a higher objective value than the ground truth one. This effect is a combination of \textit{Geometry Term} providing noisy scores for partial scans of visually similar objects (like the chairs around the table), and their relative spatial proximity, which makes the \textit{Hysteresis Term} a poor discriminator.



\section{Conclusion}
This paper presents an algorithm for estimating the semantic instance segmentation for an RGBD scan of an indoor environment.  The proposed algorithms is inductive -- using a temporal scene model which subsumes previous observations, an instance segmentation of the novel observation is inferred and used to update the temporal model.  Our experiments show better performance on a novel benchmark dataset in comparison to a strong baseline.  Interesting directions for future work include inferring the segmentation at $t_0$, investigating RNN architectures (when larger datasets become available), and replacing terms of the objective function with learned alternatives.

\section*{Acknowledgments}
\label{sec:acknowledgments}

We would like to thank Angel X. Chang and Manolis Savva for insightful discussions.  We also thank Graham et
al. \cite{3DSemanticSegmentationWithSubmanifoldSparseConvNet} and Liu et al. \cite{Liu2019MASCMA} for the comparison codes, and Dai et al. for the ScanNet data \cite{dai2017scannet}.  The project was partially supported by funding from the NSF (CRI 1729971 and VEC 1539014/1539099).

{\small
\bibliographystyle{ieee_fullname}
\bibliography{rescan}
}

\end{document}